\documentclass[runningheads]{llncs}
\usepackage{graphicx}
\usepackage{cite}
\usepackage{amsmath,amssymb,amsfonts}
\usepackage{algorithmic}
\usepackage{graphicx}
\usepackage{textcomp}
\usepackage{adjustbox}
\usepackage{multirow}
\usepackage{bbold}
\usepackage{siunitx}
\usepackage{hyperref}       
\usepackage{xcolor}
\usepackage{sidecap}
\hypersetup{
    colorlinks,
    linkcolor={blue!50!black},
    citecolor={blue!50!black},
    urlcolor={blue!50!black}
}
\usepackage{amsmath, bm}
\usepackage{hyperref}

\usepackage{amssymb}
\usepackage{pifont}

\usepackage{bbding}

\usepackage{floatrow}
\newfloatcommand{capbtabbox}{table}[][\FBwidth]

\begin{document}
\title{Rethinking Surgical Captioning: End-to-End Window-Based MLP Transformer Using Patches} 
\titlerunning{End-to-End Window-Based MLP Transformer Using Patches}

\author{Mengya Xu\inst{1,2,3*} \and
Mobarakol Islam\inst{4}\thanks{Equal technical contribution.}\and
Hongliang Ren\inst{1,2,3}\thanks{Corresponding author.}}

\authorrunning{M. Xu et al.}
%
\institute{Dept. of Electronic Engineering, The Chinese University of Hong Kong, Hong Kong SAR, China\\
\and Dept. of Biomedical Engineering, National University of Singapore, Singapore\\
\and National University of Singapore (Suzhou) Research Institute (NUSRI), China \\
\and BioMedIA Group, Dept. of Computing, Imperial College London, London, UK 
\email{mengya@u.nus.edu, m.islam20@imperial.ac.uk, hlren@ee.cuhk.edu.hk}}
\maketitle              
\begin{abstract}
Surgical captioning plays an important role in surgical instruction prediction and report generation. However, the majority of captioning models still rely on the heavy computational object detector or feature extractor to extract regional features. In addition, the detection model requires additional bounding box annotation which is costly and needs skilled annotators. These lead to inference delay and limit the captioning model to deploy in real-time robotic surgery. For this purpose, we design an end-to-end detector and feature extractor-free captioning model by utilizing the patch-based shifted window technique. We propose \textbf{S}hifted \textbf{Win}dow-Based \textbf{M}ulti-\textbf{L}ayer \textbf{P}erceptrons \textbf{Tran}sformer \textbf{Cap}tioning model (SwinMLP-TranCAP) with faster inference speed and less computation. SwinMLP-TranCAP replaces the multi-head attention module with window-based multi-head MLP. Such deployments primarily focus on image understanding tasks, but very few works investigate the caption generation task. SwinMLP-TranCAP is also extended into a video version for video captioning tasks using 3D patches and windows. Compared with previous detector-based or feature extractor-based models, our models greatly simplify the architecture design while maintaining performance on two surgical datasets. The code is publicly available at~\url{https://github.com/XuMengyaAmy/SwinMLP_TranCAP}.

\end{abstract}

\section{Introduction}
Automatic surgical captioning is a prerequisite for intra-operative context-aware surgical instruction prediction and post-operative surgical report generation. Despite the impressive performance, most approaches require heavy computational resources for the surgical captioning task, which limits the real-time deployment. The main-stream captioning models contain an expensive pipeline of detection and feature extraction modules before captioning. For example, Meshed-Memory Transformer~\cite{cornia2020meshed}, self-sequence captioning~\cite{rennie2017self} entail bounding box from detection model (Faster R-CNN~\cite{ren2015faster}) to extract object feature using a feature extractor of ResNet-101~\cite{he2016deep}. On the other hand, bounding box annotations are used to extract regional features for surgical report generation~\cite{xu2021class,xu2021learning}. However, these kinds of approaches arise following issues: (i) require bounding box annotation which is challenging in the medical scene as it requires the use of professional annotators, (ii) region detection operation leads to high computational demand, and (iii) cropping regions may ignore some crucial background information and destroy the spatial correlation among the objects. Thus recent vision-language studies~\cite{zhang2021surgical} are moving toward the detector-free trend by only utilizing image representations from the feature extractor. Nonetheless, the feature extractor still exists as an intermediate module which unavoidably leads to inadequate training and long inference delay at the prediction stage. These issues restrict the application for real-time deployment, especially in robotic surgery.

To achieve end-to-end captioning framework, ViTCAP model~\cite{fang2021injecting} uses the Vision Transformer (ViT)~\cite{dosovitskiy2020image} which encodes image patches as grid representations. However, it is very computing intensive even for a reasonably large-sized image because the model has to compute the self-attention for a given patch with all the other patches in the input image. Most recently, the window-based model Swin Transformer~\cite{liu2021Swin} outperforms ViT~\cite{dosovitskiy2020image} and reduces the computation cost by performing local self-attention between patches within the window. The window is also shifted to achieve information sharing across various spatial locations. However, this kind of approach is yet to explore for captioning tasks. 

So far, many Transformer-variants are still based on the common belief that the attention-based token mixer module contributes most to their performance. However, recent studies~\cite{tolstikhin2021mlp} show that the attention-based module in Transformer can be replaced by spatial multi-layer perceptron (MLPs) and the resulting models still obtain quite good performance. This observation suggests that the general architecture of the transformer, instead of the specific attention-based token mixer module, is more crucial to the success of the model. Based on this hypothesis,~\cite{yu2021metaformer} replaces the attention-based module with the extremely simple spatial pooling operator and surprisingly finds that the model achieves competitive performance. However, such exploration is mainly concentrated on the image understanding tasks and remains less investigated for the caption generation task.

\textbf{Our contributions can be summed up as the following points:}
1) We design SwinMLP-TranCAP, a detector and feature extractor-free captioning models by utilizing the shifted window-based MLP as the backbone of vision encoder and a Transformer-like decoder; 2) We also further develop a Video SwinMLP-TranCAP by using 3D patch embedding and windowing to achieve the video captioning task; 3) Our method is validated on publicly available two captioning datasets and obtained superior performance in both computational speed and caption generation over conventional approaches; 4) Our findings suggest that the captioning performance mostly relies on transformer type architecture instead of self-attention mechanism.

\section{Methodology}
\subsection{Preliminaries}
\subsubsection{Vision Transformer}
Vision Transformer (ViT)~\cite{dosovitskiy2020image} divides the input image into several non-overlapping patches with a patch size of $16\times16$. The feature dimension of each patch is $16\times16\times3=768$. We compute the self-attention for a given patch with all the other patches in the input image. This becomes very compute-intensive even for a reasonably large-sized image: $\Omega (MSA) = 4hwC^2 + 2 (hw)^2 $, where hw indicates the number of patches, C stands for the embedding dimension.

\begin{figure}[!t]
\centering
\includegraphics[width=0.9\linewidth]{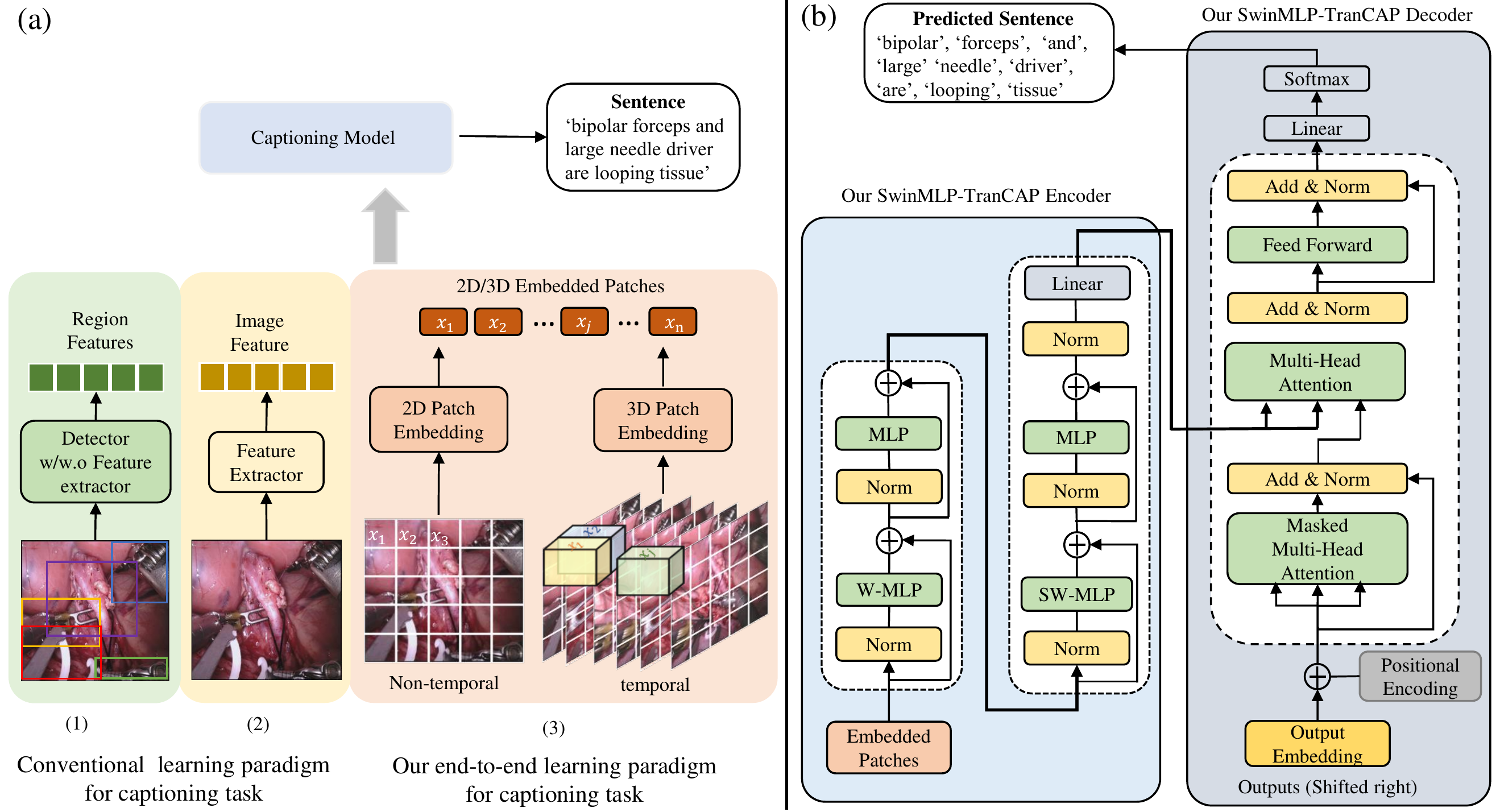}
\caption{(a) Comparisons of conventional learning paradigm and our learning paradigm for captioning task and (b) our proposed end-to-end SwinMLP-TranCAP model. (1) Captioning models based on an object detector w/w.o feature extractor to extract region features. (2) To eliminate the detector, the feature extractor can be applied as a compromise to the output image feature. (c) To eliminate the detector and feature extractor, the captioning models can be designed to take the patches as the input representation directly.}
\label{fig:model}
\end{figure}

\begin{figure}[!h]
\centering
\includegraphics[width=0.9\linewidth]{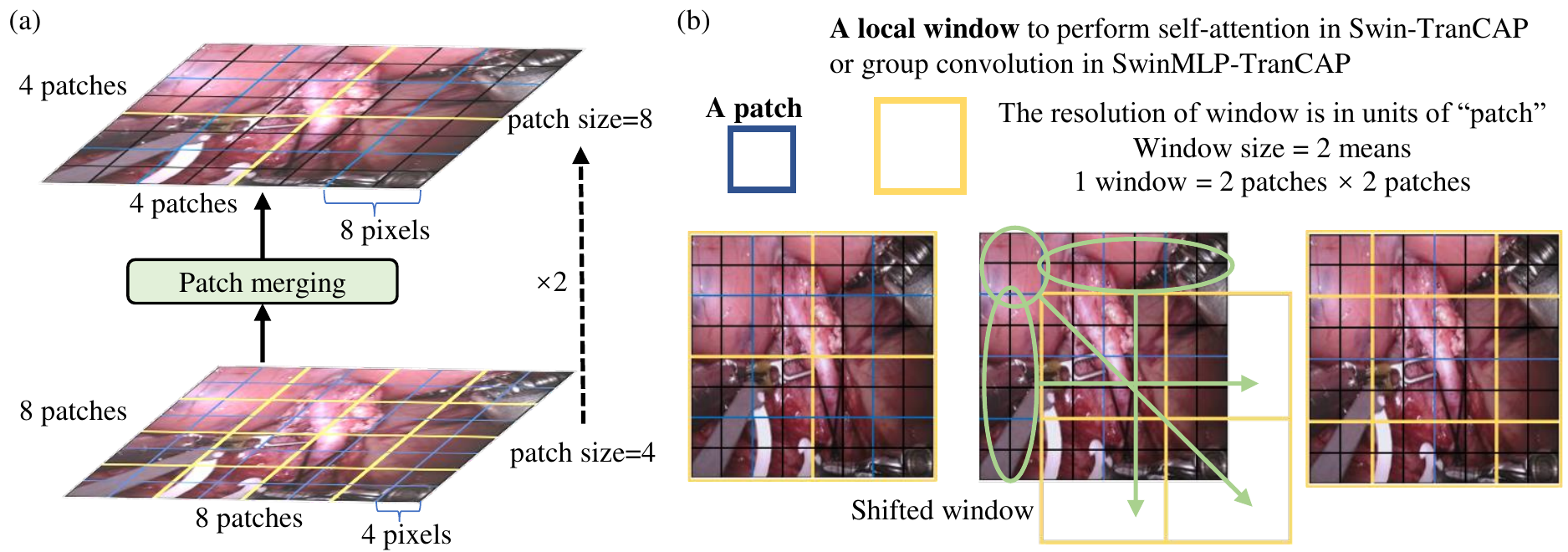}
\caption{(a) Patch merging layer in our Swin-TranCAP and SwinMLP-TranCAP. (b) shifted window to perform self-attention in Swin-TranCAP and group convolution in SwinMLP-TranCAP. For space without any pixels, cycle shifting is utilized to copy over the patches on top to the bottom, from left to right, and also diagonally across to make up for the missing patches.}
\label{fig:swin}
\end{figure}

\subsubsection{Swin Transformer}
The drawback of Multi-Head Self Attention (MSA) in ViT is attention calculation is very compute-heavy. To overcome it, Swin Transformer~\cite{liu2021Swin} introduces ``window" to perform local self-attention computation. Specifically, a single layer of transformer is replaced by two layers which design Window-based MSA (W-MSA) and Shifted Window-based MSA (SW-MSA) respectively. In the first layer with W-MSA, The input image is divided into several windows and we compute self-attention between patches within that window. The intuition is local neighborhood patches are more important than attending patches that are far away. In the second layer with SW-MSA, the shifted window is designed to achieve information sharing across various spatial locations. The computation cost can be formulated as $\Omega (\text{W-MSA}) = 4hwC^2 + 2M^2hwC$, where M stands for window size and $M^2$ indicates the number of patches inside the window.

\subsection{Our Model}
We propose the end-to-end SwinMLP-TranCAP model for captioning tasks, as shown in Fig.~\ref{fig:model} (b), which takes the embedded patches as the input representation directly. SwinMLP-TranCAP consists of a hierarchical fully multi-layer perceptron (MLP) architecture-based visual encoder and a Transformer-like decoder. Our model creates a new end-to-end learning paradigm for captioning tasks by using patches to get rid of the intermediate modules such as detector and feature extractor (see Fig.~\ref{fig:model} (a)), reduces the training parameters, and improves the inference speed. In addition, we designed the temporal version of SwinMLP-TranCAP, named Video SwinMLP-TranCAP to implement video captioning.

\textbf{2D/3D Patch Embedding Layer}
The raw image $[3, H, W]$ is partitioned and embedded into N discrete non-overlapping patches $[N, C]$ with the embedding dimension $C$ by using a 2D patch embedding layer which consists of a Conv2d projection layer (See Equation~\ref{equ:patch_embedding}) followed by flattening and transpose operation and LayerNorm. The size of each patch is $[p, p, 3]$ and the number of patches $N$ is $\frac{H}{p} \times \frac{W}{p}$. The raw feature dimension $p\times p \times 3$ of each patch will be projected into an arbitrary embedding dimension $C$. Thus, the output size from the Conv2d projection layer is $[C, \frac{H}{p}, \frac{W}{p}]$. Next, it is flattened and transposed into embedded patches of $[N, C]$. Eventually, LayerNorm is applied to embedded patches. Similarly, in the 3D patch embedding layer, the video clip $[3, T, H, W]$ is partitioned and embedded into $N$ 3D patches of $[C, \frac{T}{t}, \frac{H}{p}, \frac{W}{p}]$ (t stands for the sequence length of a cubic patch, T stands for the sequence length of a video clip) via a Conv3d projection layer.

\begin{equation}
Conv2d(in=3, out=C, kernel\_size=p, stride=p)
\label{equ:patch_embedding}
\end{equation}

\textbf{Swin-TranCAP}
We utilize Swin Transformer~\cite{liu2021Swin} as the backbone of the vision encoder and Transformer-like decoder to implement the window-based self-attention computation and reduce the computation cost.

\textbf{SwinMLP-TranCAP}
SwinMLP-TranCAP consist of a SwinMLP encoder and a Transformer-like decoder, as shown in Fig.~\ref{fig:swin} (b). Embedded patches $[\frac{H}{p} \times \frac{W}{p}, C]$ are used as the input representation and each patch is treated as a ``token". In SwinMLP encoder, we replace the Window/Shifted-Window MSA with Window/Shifted-Window Multi-Head Spatial MLP (W-MLP and SW-MLP) with the number of heads $n$ and window size $M$ via group convolution for less-expensive computation cost, which can be formulated as

\begin{equation}
Conv1d(in= nM^2, out=nM^2, kernel\_size=1, groups= n )
\end{equation}

In two successive Swin MLP blocks, the first block with MLP is applied to image patches within the window independently and the second block with MLP is applied across patches by shifting window. The vision encoder consists of $4$ stages and a linear layer. The Swin MLP block in the first stage maintain the feature representation as $[\frac{H}{p} \times \frac{W}{p}, C]$. Each stage in the remaining $3$ stages contains a patch merging layer and multiple ($2 \times$) Swin MLP Block. The patch merging layer (see Fig.~\ref{fig:swin} (a)) concatenates the features of each group of $2\times 2$ neighboring patches and applies a linear layer to the $4C$ dimensional concatenated features. This decreases the amount of ``tokens" by $2\times 2 = 4$, and the output dimension is set to $2C$. Thus the feature representation after patch merging layer is $[\frac{H}{2 \times p} \times \frac{W}{2 \times p}, 2C]$. Afterward, Swin MLP blocks maintain such a shape of feature representation. Swin MLP blocks are created by replacing the standard MSA in a Transformer block with the window-based multi-head MLP module while keeping the other layer the same. The block in the vision encoder consists of a window-based MLP module, a 2-layer MLP with GELU nonlinearity. Before each windowed-based MLP module and each MLP, a LayerNorm layer is applied, and a residual connection is applied after each module. Overall, the number of tokens is reduced and the feature dimension is increased by patch merging layers as the network gets deeper. The output of 4 stages is $[\frac{H}{2^3 \times p} \times \frac{W}{2^3 \times p},  2^3 \times C ]$ (3 means the remaining 3 stages). Eventually, a linear layer is applied to produce the $[\frac{H}{2^3 \times p} \times \frac{W}{2^3 \times p}, 512]$ feature representation to adapt the Transformer-like decoder. The decoder is a stack of $6$ identical blocks. Each block consists of two multi-head attention layers (an encoder-decoder cross attention layer and a decoder masked self-attention layers) and a feed-forward network.

\textbf{Video SwinMLP-TranCAP}
We form the video clip $x=\left\{x_{t-T+1}, ..., x_t\right\}$ consisting the current frame and the preceding $T-1$ frames. The video clip is partitioned and embedded into $\frac{T}{t} \times \frac{H}{p} \times \frac{W}{p}$ 3D patches and each 3D patch has a $C$-dimensional feature via a 3D Patch Embedding layer. Each 3D patch of size $[t, p, p]$ is treated as a ``token". The major part is the video SwinMLP block which is built by the 3D shifted window-based multi-head MLP module while keeping the other components same. Given a 3D window of $[P, M, M]$, in two consecutive layers, the multi-head MLP module in the first layer employs the regular window partition approach to create non-overlapping 3D windows of $[\frac{T}{t\times P} \times \frac{H}{p \times M} \times \frac{W}{p \times M}]$. In the second layer with multi-head MLP module, the window is shifted along the temporal, height, and width axes by $[\frac{P}{2}, \frac{M}{2}, \frac{M}{2}]$ tokens from the first layer's output.

\textbf{Our models depart from ViT~\cite{dosovitskiy2020image} and Swin Transformer~\cite{liu2021Swin} with several points:}
1) no extra ``class" token;
2) no self-attention blocks in the visual encoder: it is replaced by MLP via group convolutions;
3) Designed for captioning tasks. The window-based visual encoder is paired with a Transformer-like decoder via a linear layer which reduces the dimensional to $512$ to adapt to the language decoder;
4) use window size of $14$, instead of $7$ used in Swin Transformer~\cite{liu2021Swin}.

\section{Experiments}
\subsection{Dataset Description} 

\textbf{DAISI dataset} The Database for AI Surgical Instruction (DAISI)~\cite{rojas2020daisi} \footnote{\url{https://engineering.purdue.edu/starproj/_daisi/}} contains $17339$ color images of $290$ different medical procedures, such as laparoscopic sleeve gastrectomy, laparoscopic ventral hernia repair, tracheostomy, open cricothyroidotomy, inguinal hernia repair, external fetal monitoring, IVC ultrasound, etc. We use the filtered DAISI dataset cleaned by~\cite{zhang2021surgical} which removes noisy and irrelevant images and text descriptions. We split the DAISI dataset into $13094$ images for training, and $1646$ images for validation by following~\cite{zhang2021surgical}.

\textbf{EndoVis 2018 Dataset} is from the MICCAI robotic instrument segmentation dataset\footnote{\url{https://endovissub2018-roboticscenesegmentation.grand-challenge.org/}} of endoscopic vision challenge 2018 \cite{allan20202018}. The training set consists of $15$ robotic nephrectomy procedures acquired by the da Vinci X or Xi system. We use the annotated caption generated by ~\cite{xu2021class} which employs $14$ sequences out of the $15$ sequences while disregarding the 13th sequence due to the less interaction. The validation set consists of the 1st, 5th, and 16th sequences, and the train set includes the 11 remaining sequences following the work\cite{xu2021class}.

\subsection{Implementation details}
Our models are trained using cross-entropy loss with a batch size of $9$ for $100$ epochs. We employ the Adam optimizer with an initial learning rate of $3e-4$ and follow the learning rate scheduling strategy with $20000$ warm-up steps. Our models are implemented on top of state-of-the-art captioning architecture, Transformer~\cite{dosovitskiy2020image}. The vanilla Transformer architecture is realized with the implementation~\footnote{\url{https://github.com/ruotianluo/ImageCaptioning.pytorch}} from ~\cite{dosovitskiy2020image}. All models are developed with Pytorch and trained with NVIDIA RTX3090 GPU. In our models, the embedding dimension $C=128$, patch size $p=4$, and window size $M=14$. In Video SwinMLP-TranCAP, we use the sequence length $T$ of $4$ and 3D patch of $[2, 4, 4]$ and window size of $[2, 14, 14]$.

\section{Results and Evaluation}

\begin{table}[]
\centering
\caption{Comparison of hybrid models and our models. 
The DAISI dataset does not have object annotation and video captioning annotation. Thus the hybrid of YLv5 w. RN18 and Transformer, and Video SwinMLP-Tran experiments are left blank.}
 \scalebox{0.8}{
    \begin{tabular}{c|cc|c|c|c|c|cccc}
    \hline
    \multicolumn{3}{c|}{Model} & \multicolumn{4}{c|}{DAISI Dataset} & \multicolumn{4}{c}{EndoVis18 Dataset}\\
    \hline
    \multicolumn{1}{c}{Det} & \multicolumn{1}{c}{FE} & \multicolumn{1}{c|}{Captioning Model} & \multicolumn{1}{c}{B4} & \multicolumn{1}{c}{MET} & \multicolumn{1}{c}{SPI} & CID   & B4    & MET   & SPI   & CID \\
    \hline
    \multicolumn{1}{c}{\multirow{3}[2]{*}{FasterRCNN~\cite{ren2015faster}}} & \multirow{3}[2]{*}{RN18~\cite{he2016deep}} & Tran~\cite{dosovitskiy2020image}  & \multicolumn{4}{c|}{\multirow{3}[2]{*}{\XSolidBrush}} &  0.363 & 0.323 & 0.512 & 2.017 \\
    \multicolumn{1}{c}{} &       & Self-Seq~\cite{rennie2017self} & \multicolumn{4}{c|}{}         & 0.295 & 0.283 & 0.496 & 1.801 \\
    \multicolumn{1}{c}{} &       & AOA~\cite{huang2019attention}   & \multicolumn{4}{c|}{}         &  0.377 & 0.371 & 0.58  & 1.811 \\
    \hline
    \multicolumn{1}{c}{YLv5x~\cite{glenn_jocher_2022_6222936}} & RN18~\cite{he2016deep}  & Tran~\cite{dosovitskiy2020image}  & \multicolumn{4}{c|}{\XSolidBrush}        & 0.427 & 0.328 & 0.577 & \textbf{3.022} \\
\cline{1-7}    \multicolumn{1}{c}{\multirow{3}[2]{*}{\XSolidBrush}} & \multirow{3}[2]{*}{RN18~\cite{he2016deep}} & Self-Seq~\cite{rennie2017self} & \multicolumn{1}{c}{0.296} & \multicolumn{1}{c}{0.207} & \multicolumn{1}{c}{0.330} & 2.827 & 0.446 & 0.353 & 0.531 & 2.674 \\
    \multicolumn{1}{c}{} &       & AOA~\cite{huang2019attention}   & \multicolumn{1}{c}{0.349} & \multicolumn{1}{c}{0.246} & \multicolumn{1}{c}{0.403} & 3.373 & 0.427 & 0.322 & 0.533 & 2.903 \\
    \multicolumn{1}{c}{} &       & Tran~\cite{zhang2021surgical}  & \multicolumn{1}{c}{0.454} & \multicolumn{1}{c}{0.308} & \multicolumn{1}{c}{\textbf{0.479}} & \textbf{4.283} & 0.426 & 0.335 & 0.524 & 2.826\\
    \hline
    \multicolumn{1}{c}{\XSolidBrush} & 3DRN18 & Tran~\cite{dosovitskiy2020image}  & \multicolumn{4}{c|}{\XSolidBrush}        &  0.406 & 0.345 & 0.586 &  2.757\\
    \hline
    \multicolumn{2}{c}{\multirow{3}[2]{*}{Ours}} & Swin-TranCAP & \multicolumn{1}{c}{0.346} & \multicolumn{1}{c}{0.237} & \multicolumn{1}{c}{0.378} & 3.280 & \textbf{0.459} & 0.336 & 0.571 & 3.002 \\
    \multicolumn{2}{c}{} & SwinMLP-TranCAP & \multicolumn{1}{c}{\textbf{0.459}} & \multicolumn{1}{c}{\textbf{0.308}} & \multicolumn{1}{c}{0.478} & 4.272 & 0.403 & 0.313 & 0.547 & 2.504 \\
    \multicolumn{2}{c}{} & V-SwinMLP-TranCAP & \multicolumn{4}{c|}{\XSolidBrush}        & 0.423 & \textbf{0.378} & \textbf{0.619} & 2.663 \\
    \hline
    \end{tabular}
}
\label{tab_main}
\end{table}

We evaluate the captioning models using the BLEU-4 (B4)~\cite{papineni2002bleu}, METEOR (MET)~\cite{banerjee2005meteor}, SPICE (SPI)~\cite{anderson2016spice}, and CIDEr (CID)~\cite{vedantam2015cider}. We employ the ResNet18~\cite{he2016deep} pre-trained on the ImageNet dataset as the feature extractor (FE), YOLOv5 (YLv5)~\cite{glenn_jocher_2022_6222936}, and FasterRCNN~\cite{ren2015faster} pre-trained on the COCO dataset as the detector (Det). The captioning models includes self-sequence captioning model (Self-Seq)~\cite{rennie2017self}, Attention on Attention (AOA) model~\cite{huang2019attention}, Transformer model~\cite{dosovitskiy2020image}. Self-sequence and AOA originally take the region features extracted from the object detector with feature extractor as input. In our work, we design the hybrid style for them by sending image features extracted by the feature extractor only. 3D ResNet18 is employed to implement the video captioning task. We define the end-to-end captioning models take the image patches directly as input, trained from scratch as \textbf{Ours}, including Swin-TranCAP and SwinMLP-TranCAP for image captioning and Video SwinMLP-TranCAP (V-SwinMLP-TranCAP) for video captioning.

\begin{table}[]
\centering
\caption{Proof of ``less computation cost" of our approaches with FPS, Num of Parameters, and GFLOPs.}
\scalebox{0.8}{
\begin{tabular}{ccc|ccc}
    \hline
    \multicolumn{3}{c|}{Model} & \multicolumn{3}{c}{Proof of less-computation cost} \\
    \hline
    Det  & \multicolumn{1}{c}{FE} & \multicolumn{1}{c|}{Captioning Model} & FPS   & N\_Parameters(M) & GFLOPs\\
    \hline
    FasterRCNN~\cite{ren2015faster} & RN18~\cite{he2016deep}  & Tran~\cite{dosovitskiy2020image}  &  8.418 & 28.32+46.67 &  251.84+25.88 \\
    \hline
    YLv5x~\cite{glenn_jocher_2022_6222936} & RN18~\cite{he2016deep}  & Tran~\cite{dosovitskiy2020image}  & 9.368 & 97.88+46.67 & 1412.8+25.88 \\
    \hline
    \XSolidBrush     & RN18~\cite{he2016deep}  & Tran~\cite{zhang2021surgical}  &  11.083 & 11.69+46.67 & 1.82+25.88 \\
    \hline
    \multicolumn{2}{c}{\multirow{2}[2]{*}{Ours}} & Swin-TranCAP & 10.604 & 165.51 &  19.59 \\
    \multicolumn{2}{c}{} & SwinMLP-TranCAP & \textbf{12.107} & \textbf{99.11} & \textbf{14.15}\\
    \hline
    \end{tabular}}
\label{tab_real_time_proof}
\end{table}

\begin{figure}[!h]
\centering
\includegraphics[width=0.9\linewidth]{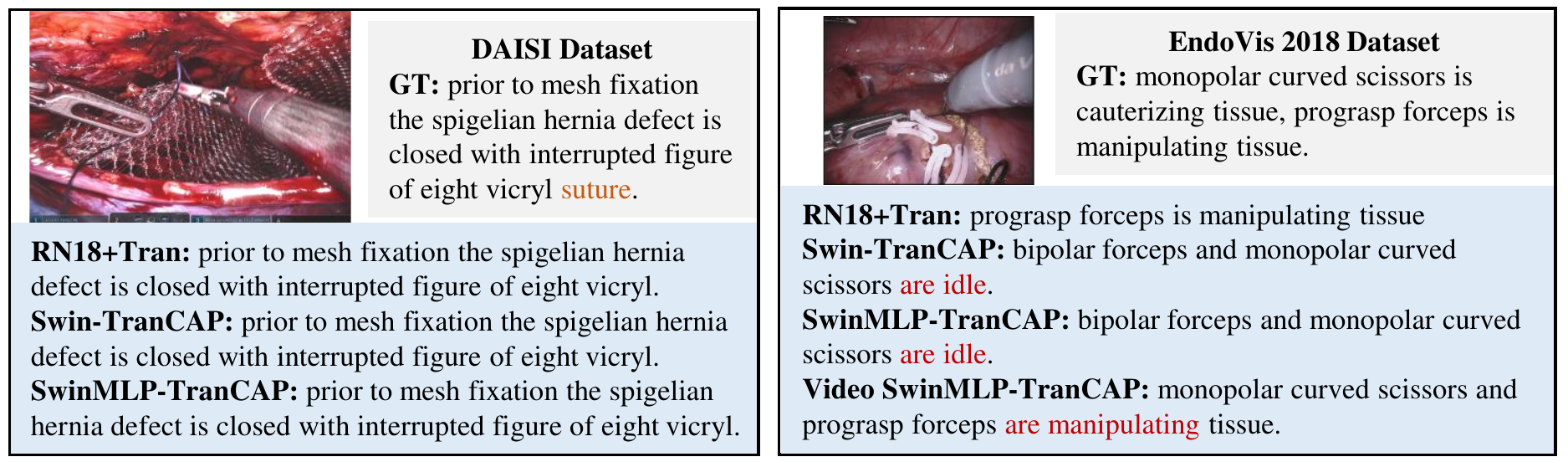}
\caption{Qualitative results with our model and hybrid models.}
\label{fig:sentence}
\end{figure}

We now briefly describe how to use the detector and feature extractor extract features for captioning model. The spatial features from YOLOv5 w. ResNet are $(X, 512)$. $X$ indicates the number of predicted regions, and $X$ varies from image to image. The first $N$ predicted regions are sent to the captioning model. Zero appending is used to deal with those images where $N>X$ ($N=6$ in our work). When only using the feature extractor, we take the 2D adaptive average pooling of the final convolutional layer output, which results in $14 \times 14 \times 512$ output for each image. It is reshaped into $196\times512$ before sending into the captioning model. We find that our models can achieve decent performance even without the need for a detector and feature extractor (see Table~\ref{tab_main} and Fig.~\ref{fig:sentence}). On the DAISI dataset, our SwinMLP-TranCAP preserves the performance compared with the hybrid Transformer model and outperforms the FC and AOA model including $11\%$ gain in BlEU-4. On EndoVis 2018 Dataset, our Swin-TranCAP model shows $4.7\%$ improvement in SPICE compared with the hybrid Transformer model. Although the performance of SwinMLP-TranCAP drops a bit, it has less computation. In Table~\ref{tab_real_time_proof}, less computation cost of our approaches is proven by evaluating FPS, Num of Parameters, and GFLOPs. Our approach achieves better captioning performance/efficiency trade-offs. It is worth highlighting that our purpose is not to provide better results but to remove the object detector and feature extractor from the conventional captioning system training pipeline for more flexible training, less computation cost, and faster inference speed without sacrificing performance. Surprisingly, our approach also obtained a slightly better quantitative performance.

\begin{table*}
\begin{floatrow}
\scalebox{0.9}{
\capbtabbox{

\begin{tabular}{c|cc|cc}
\hline
\multicolumn{1}{l|}{} & \multicolumn{2}{c|}{DAISI}                           & \multicolumn{2}{c}{EndoVis18}                     \\ \hline
p            & \multicolumn{1}{c}{B4}         & CID         & \multicolumn{1}{c}{B4}         & CID         \\ \hline
4                     & \multicolumn{1}{c}{0.192}          & 1.703          & \multicolumn{1}{c}{0.339}          & 2.105          \\ 
8                     & \multicolumn{1}{c}{0.247}          & 2.195          & \multicolumn{1}{c}{0.364}          & 1.886          \\ 
16                    & \multicolumn{1}{c}{\textbf{0.302}} & \textbf{2.776} & \multicolumn{1}{c}{\textbf{0.416}} & \textbf{3.037} \\ \hline
\end{tabular}
}{
 \caption{Tune patch size $p$ for vanilla Transformer.}
 \label{tab:vit_cap}
}
\capbtabbox{
\begin{tabular}{c|cc|cc|cc}
\hline
               & \multicolumn{2}{c|}{Para.}                & \multicolumn{2}{c|}{DAISI}         & \multicolumn{2}{c}{EndoVis18}  \\ \hline
Model          & \multicolumn{1}{c|}{C}   & M                   & \multicolumn{1}{c}{B4}    & CID   & \multicolumn{1}{c}{B4}    & CID   \\ \hline
Swin-Tran-S    & \multicolumn{1}{c|}{96}  & \multirow{4}{*}{14} & \multicolumn{1}{c}{0.329} & 3.109 & \multicolumn{1}{c}{\textbf{0.471}} & \textbf{3.059} \\ \cline{1-2} \cline{4-7} 
Swin-Tran-L    & \multicolumn{1}{c|}{128} &                     & \multicolumn{1}{c}{0.346} & 3.280 & \multicolumn{1}{c}{0.459} & 3.002 \\ \cline{1-2} \cline{4-7} 
SwinMLP-Tran-S & \multicolumn{1}{c|}{96}  &                     & \multicolumn{1}{c}{0.455} & 4.188 & \multicolumn{1}{c}{0.398} & 2.322 \\ \cline{1-2} \cline{4-7} 
SwinMLP-Tran-L & \multicolumn{1}{c|}{128} &                     & \multicolumn{1}{c}{\textbf{0.459}} & \textbf{4.272} & \multicolumn{1}{c}{0.403} & 2.504 \\ \hline
Swin-Tran-L    & \multicolumn{1}{c|}{128} & 7                   & \multicolumn{1}{c}{0.433} & 4.049 & \multicolumn{1}{c}{0.389} & 2.932 \\ \hline
SwinMLP-Tran-L & \multicolumn{1}{c|}{128} & 7                   & \multicolumn{1}{c}{0.434} & 4.046 & \multicolumn{1}{c}{0.403} & 2.707 \\ \hline
\end{tabular}
}{
 \caption{Tune embedding dimension $C$ with patch size $p$ of 4 and window size $M$ of $14$, for our models.}
 \label{tab:tune_ours}
}}
\end{floatrow}
\end{table*}

\section{Ablation Study}
We simply incorporate the patch embedding layer into the vanilla Transformer captioning model and report the performance in Table~\ref{tab:vit_cap}. The embedding dimension $C$ is set to $512$. And we also investigate the performance of different configurations for our models (see Table~\ref{tab:tune_ours}). $C=96$ paired with layer number $(2, 2, 6, 2)$ is \textbf{S}mall version and $C=128$ paired with layer number $(2, 2, 18, 2)$ is \textbf{L}arger version. We find that the large version with a window size of $14$ performs slightly better.

\section{Discussion and Conclusion}
We present the end-to-end SwinMLP-TranCAP captioning model, and take the image patches directly, to eliminate the object detector and feature extractor for real-time application. The shifted window and multi-head MLP architecture design make our model less computation. Video SwinMLP-TranCAP is also developed for video captioning tasks. Extensive evaluation of two surgical captioning datasets demonstrates that our models can maintain decent performance without needing these intermediate modules. Replacing the multi-head attention module with multi-head MLP also reveals that the generic transformer architecture is the core design, instead of the attention-based module. Future works will explore whether the attention-based module in a Transformer-like decoder is necessary or not. Different ways to implement patch embedding in the vision encoder are also worth studying.

\subsubsection{Acknowledgements.}
This work was supported by the Shun Hing Institute of Advanced Engineering (SHIAE project BME-p1-21) at the Chinese University of Hong Kong (CUHK), Hong Kong Research Grants Council (RGC) Collaborative Research Fund (CRF C4026-21GF and CRF C4063-18G) and (GRS)\#3110167.

\bibliography{mybib}{}
\bibliographystyle{splncs04}
\end{document}